\documentclass[a4paper]{article}
\usepackage{geometry}
\usepackage[utf8]{inputenc}
\usepackage{hyperref}
\usepackage{geometry}
\usepackage{comment}
\usepackage{cite}
\usepackage{amsmath,amssymb,amsfonts}
\usepackage{algorithmic}
\usepackage{graphicx}
\usepackage{textcomp}
\usepackage{xcolor}
\usepackage{fourier}
\usepackage{float}
\usepackage{caption}
\usepackage{subcaption}
\usepackage{url}
\usepackage{color}
\usepackage{mathrsfs}
\usepackage{array}

\DeclareMathOperator*{\argmax}{arg\,max}

\date{}

\providecommand{\keywords}[1]
{
  \small	
  \textbf{\textit{Keywords---}} #1
}

\title{Impact of Spatial Frequency Based Constraints on Adversarial Robustness}

\author{Rémi Bernhard\textsuperscript{1,2}, Pierre-Alain Moellic\textsuperscript{1,2}, Martial Mermillod\textsuperscript{3}, Yannick Bourrier\textsuperscript{3},\\ Romain Cohendet\textsuperscript{4}, Miguel Solinas\textsuperscript{4}, Marina Reyboz\textsuperscript{4}\\ \\
\small \textsuperscript{1}CEA Tech, Centre CMP, Equipe Commune CEA Tech - Mines Saint-Etienne, F-13541 Gardanne, France\\
\small \textsuperscript{2}Univ. Grenoble Alpes, CEA, Leti, F-38000 Grenoble, France\\ 
\small \textit{\{remi.bernhard, pierre-alain.moellic\}@cea.fr}\\
\small\textsuperscript{3} LPNC, CNRS, Université Grenoble Alpes, Université Savoie Mont Blanc, Grenoble, France\\
\small \textit{ \{martial.mermillod, yannick.bourrier\}@univ-grenoble-alpes.fr}\\
\small\textsuperscript{4} Univ. Grenoble Alpes, CEA, List, F-38000 Grenoble, France\\
\small \textit{ \{romain.cohendet, miguelangel.solinas, marina.reyboz\}@cea.fr}\\
}

\begin{document}

\maketitle

\begin{abstract}
Adversarial examples mainly exploit changes to input pixels to which humans are not sensitive to, and arise from the fact that models make decisions based on uninterpretable features. 
Interestingly, cognitive science reports that the process of interpretability for human classification decision relies predominantly on low spatial frequency components. In this paper, we investigate the robustness to adversarial perturbations of models enforced during training to leverage information corresponding to different spatial frequency ranges. We show that it is tightly linked to the spatial frequency characteristics of the data at stake. Indeed, depending on the data set, the same constraint may results in very different level of robustness (up to $0.41$ adversarial accuracy difference). To explain this phenomenon, we conduct several experiments to enlighten influential factors such as the level of sensitivity to high frequencies, and the transferability of adversarial perturbations between original and low-pass filtered inputs.
\end{abstract} \hspace{10pt}


\section{Introduction}
\label{ntroduction}
Neural networks based models have been shown to reach impressive performances on challenging tasks, while being vulnerable to adversarial examples, i.e. maliciously crafted perturbations added to clean examples to fool a model at inference
~\cite{szegedy2013intriguing}. This phenomenon has shed the light on the fact that, to
perform a specific task, machine learning models rely on different features or different feature processing from those humans rely on to make their decisions \cite{jo2017measuring, ilyas2019adversarial, yinfourier2020}.

To make a model robust against an adversary in the white-box setting, many defenses have been developed including proactive \cite{Madry2017, zhang2019theoretically, hendrycks2019, cohen2019certified}. 
Few works underline the critical link between robustness and the interpretability of the features a model relies on: it has been shown that models trained with adversarial training \cite{Madry2017} or randomized smoothing \cite{cohen2019certified} exhibit interpretable gradients \cite{tsipras2018robustness, etmann2019connection, kaur2019aligned} and, inversely,  a recent effort \cite{chan2020jacobian} demonstrates that a model trained to have explainable jacobian matrices presents adversarial robustness. 

Furthermore, Zhang \textit{et al.} \cite{Zhang2019interpreting} experimentally highlight the importance of low spatial frequency (hereafter, LSF) information, such as shape, for adversarially trained models, in opposition to concepts associated with high spatial frequency (hereafter, HSF) information. 
Experimental evidence in neural computation and cognitive psychology suggests the importance of LSF to perform efficient classification~\cite{schyns1994blobs, mermillod2010coarse, french2002importance}. Therefore, a natural hypothesis would be that a model trained specifically to rely more on LSF information might present an improved adversarial robustness. This hypothesis has already been indirectly exploited by taking advantage of preprocessing defense schemes based on HSF components filtering~\cite{das2018shield, liu2019feature, zhang2019adversarial}. Other defenses aim at training a model exploiting more human interpretable information, such as \cite{addepalli2020towards} by giving higher
importance to information present in higher bit planes. However, these methods stay agnostic of the intrinsic spatial frequency characteristics of the data. 

The objective of this work is twofold. First, we experimentally question some preconceived hypothesis related to adversarial examples, more particularly ones considering adversarial perturbations as a pure HSF phenomenon with data-agnostic spatial frequency characteristics. Second, we aim at 
investigating the link between spatial frequency features of the information that a model uses to perform predictions \textit{and} the robustness against adversarial perturbations offered by spatial frequency-based constraints.  
Our key contributions are:
\begin{itemize}
    \item We show that a frequency-based regularization induces very different levels of robustness according to the frequency features of the data. As an example, a low-frequency constrained model on CIFAR10 (that covers a broad frequency spectrum) has no robustness, while reaching a $41$ \% true robustness for SVHN against the $l_{\infty}$ PGD attack.
    \item By analyzing the sensitivity of a model trained naturally (i.e. without spatial frequency-based procedures and hereafter noted as \textit{regular} model) as well as adversarial transferability properties, we observe that enforcing a model to rely on LSF information is not a necessary condition to bring adversarial robustness.
    \item We notice that, depending on the data set complexity, some models spread over the whole frequency spectrum, and show that constraints spanning different frequency ranges can help improving robustness.
    \item We discuss combination with adversarial training \cite{Madry2017} for future overall defense strategy.
\end{itemize}
The paper is organized as follows. After positioning our work in relation to the state-of-the-art in Section \ref{Related work}, 
we analyze, in Section \ref{Frequency analysis and transferability}, frequency properties and sensitivity of features learned by a model. Notably, we study to which extent features learned by a model are focused on low or high spatial frequency concepts, and the sensitivity of models to frequency constrained noise. 
In Section \ref{Transferability}, we set forth interesting links between transferability of adversarial perturbations and frequency properties of the information models make use of. In Section \ref{Constraints on frequencies}, we design loss functions to force a model to extract features in particular frequency ranges, and observe the potential effects on adversarial robustness. We link these effects with analysis performed in Sections~\ref{Frequency analysis and transferability} and~\ref{Transferability}. We also show promising future work direction by binding our findings with Adversarial Training.

\section{Related work}
\label{Related work}
Recent efforts~\cite{wang2020highfrequency, yinfourier2020} experimentally demonstrate that regular models predominantly exploit non-interpretable HSF components, and that robust models tend to use more concepts, such as shape, associated to LSF~\cite{geirhos2018imagenettrained, Zhang2019interpreting}, in a way similar to that of humans~\cite{schyns1994blobs, mermillod2010coarse, french2002importance}.
Therefore, a common belief is that the adversarial vulnerability of a model comes from the utilization of HSF components \cite{wang2020towards}. However, Yin \textit{et al.}~\cite{yinfourier2020} show that adversarial perturbations cannot be viewed only as a HSF phenomenon: adversarially trained models, more sensitive to LSF, stay vulnerable to an adversary that optimally exploits this part of the frequency spectrum. Notably, Sharma \textit{et al.}~\cite{sharma2019lowfreq} make use of LSF constraints to craft adversarial examples against adversarially trained models on ImageNet and use fewer iterations than classical adversarial attacks.

As adversarial examples are the consequence of models relying on brittle and non-interpretable features \cite{ilyas2019adversarial}, some papers exploit the idea of enforcing a model to use features to which humans are sensitive.
During training, Yin \textit{et al.} \cite{yinfourier2020} add LSF noise and demonstrate that it does not necessarily improve robustness to LSF perturbations. As a explanation, the authors hypothesize that as natural images are more LSF concentrated, it is harder for a model to become invariant to these spatial frequencies and, then, to LSF perturbations. Before, Geirhos \textit{et al.} \cite{geirhos2018imagenettrained} 
interestingly took advantage of stylized images that keep only shape information within a data augmentation process to improve the robustness against common perturbations.
Even if not related to frequency concerns, Addepalli \textit{et al.}  show in \cite{addepalli2020towards} that constraining a model to rely on the information contained in the higher bit planes only (as humans make decisions based on the information of large magnitude) has a positive impact on robustness.

\section{Preliminaries}
\label{Preliminaries}

    \subsection{Notations}
    \label{Notations}
    
A neural network model $M_\theta$, with parameters $\theta$, classifies an input $x \in \mathbb{R}^d$ to a label $M_{\theta}(x) \in \left\{ 1 \dots C \right\}$. $L^{E}(\theta, x, y)$ denotes the cross-entropy loss for $M_\theta$ and $(x,y)$ an input with its corresponding ground-truth label.
The pre-softmax function of $M_\theta$ (the logits) is denoted as $f: \mathbb{R}^d \rightarrow \mathbb{R}^C$.
We denote $x^{low}_i$ and $x^{high}_i$ respectively the low-pass and high-pass filtered versions of $x$ at intensity $i$ (see Section \ref{Filtering with the Fourier transform}). 
The \textit{LSF task} (resp. \textit{HSF task}) refers to the classification task where inputs have been low-pass (resp. high-pass) filtered at some intensity, i.e. where input-label pairs correspond to $(x^{low}_i, y)$ (resp. $(x^{high}_i, y)$) for some $i$. $M^{low}_i$ (resp. $M^{high}_i$) denotes a model trained for the LSF (resp. HSF) task.
The accuracy is denoted by $Acc$, and the adversarial accuracy ($Acc_{adv}$) denotes the accuracy of a model on a set of adversarial examples.

\begin{figure*}[t]
\begin{center}
   \includegraphics[width=0.95\linewidth]{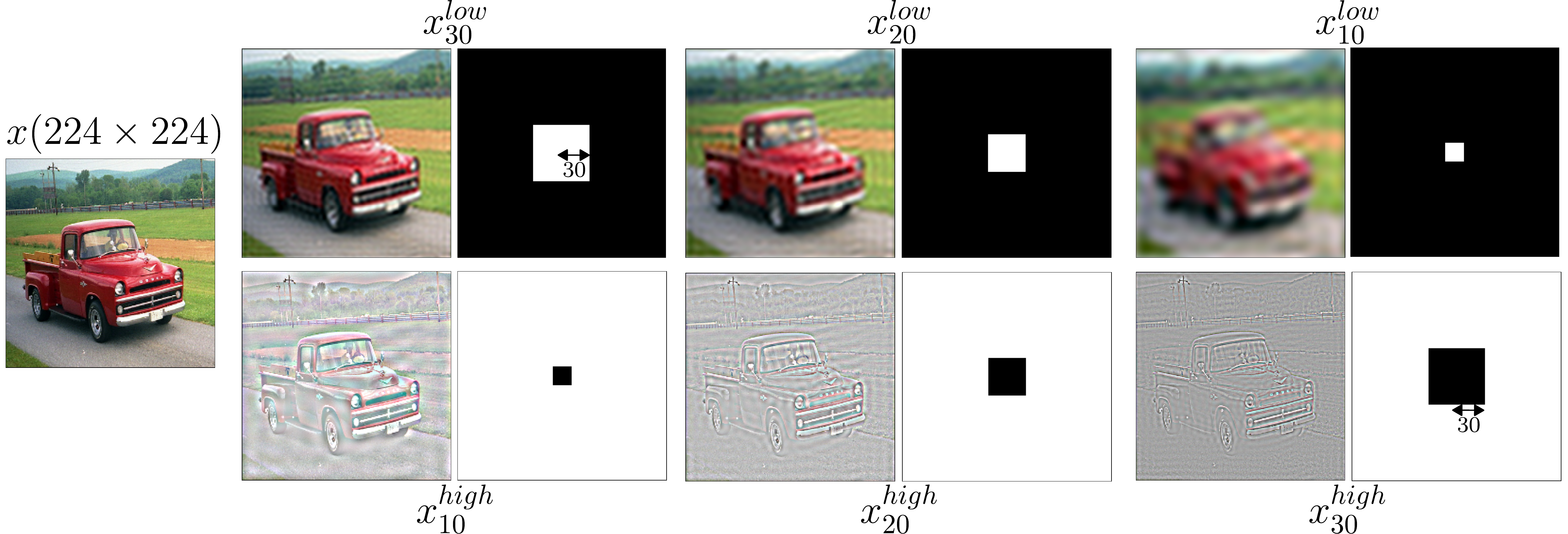}
\end{center}
\caption{Original image (left). Low-pass filtered images with corresponding mask (first row). High-pass filtered images (magnified for visualization) with corresponding mask (bottom row). For the Fourier domain masks, white denotes value $1$ and black value $0$.~\textbf{For LSF, low filtering intensity means restricted low-pass filtering. For HSF, high filtering intensity means restricted high-pass filtering.}}
\label{filt_exemple}
\end{figure*}

    \subsection{Filtering with the Fourier transform}
    \label{Filtering with the Fourier transform}

Low or high-pass filtering is performed by deleting the undesired spatial frequencies in the Fourier domain thanks to a boolean mask $\Omega \in \left\{ 0,1 \right\}^{n \times n}$, similarly to~\cite{sharma2019lowfreq, wang2020highfrequency, wang2020towards, yinfourier2020}. We note $\mathcal{F}$ and $\mathcal{F}^{-1}$ the Discrete Fourier Transform (DFT), and its inverse function, respectively. For a gray-scale image $a \in \mathbb{R}^{n \times n}$,
$\mathcal{F}_c(a) \in \mathbb{C}^{n \times n}$ denotes the centered variant of $\mathcal{F}(a) \in \mathbb{C}^{n \times n}$ (i.e. coeffients for low frequencies are located at the center, and those for high-frequencies at the corners). The filtered image is then obtained classically as $x^{freq}=\mathcal{F}^{-1}\big(\mathcal{F}_c(a) \odot \Omega\big)$, with $\odot$ the Hadamard product.  
For a color image, the procedure is applied to each of the channels.
We note $x^{low}_i$ for $\Omega$ corresponding to 1's only in the $2i \times 2i$ square in the middle, and $x^{high}_i$ for $\Omega$ corresponding to 1's only outside the $2i \times 2i$ square. In consequence, for low-pass (resp. high-pass) filtering, the smaller (resp. the higher) the intensity $i$, the stronger the filter is.
We illustrate this process in Figure~\ref{filt_exemple}.

    \subsection{Data sets and models}
    \label{data_sets_models}

We consider CIFAR10 \cite{kriz12}, SVHN \cite{Netzer2011} and a custom-built data set, named \textit{Small ImageNet}, built by extracting $10$ meta classes from the ImageNet ILSVRC2012 benchmark. For each meta class, we extracted $3000$ images from the original training set and $300$ images from the non-blacklisted validation set. All input images are scaled to $\left[ 0,1 \right]$.
CIFAR10 and SVHN, both involving color images of size $32 \times 32$ allow to make comparisons and draw conclusions about phenomenon observed. \textit{Small ImageNet}, composed of $224 \times 224$ color images, enables to extrapolate results on images with higher definition.
For SVHN we use a model inspired from VGG \cite{Simonyan15}, for CIFAR10 we use a WideResNet28-8 model \cite{Zagoruyko2016WRN}, and for \textit{Small Imagenet} we consider a MobileNetV2 model \cite{mobilenetv2}.
All details about \textit{Small Imagenet} and the training hyper-parameters are presented in the code repository of this work\footnote{\url{https://gitlab.emse.fr/remi.bernhard/Frequency_and_Robustness/}\label{gitlab}}.

\section{Frequency properties of data and models}
\label{Frequency analysis and transferability}
In this section we aim at gaining insight on the way information learned by a classifier trained to solve the regular classification task contains information for the LSF and HSF tasks.
These experiments allow to gain intuition on the way models leverage information with respect to data sets frequencies, and will help to better understand robustness dissimilarities which occur between different models trained with the same frequency-based constraints (later in Section \ref{Constraints on frequencies}).

\subsection{Impact of filtered data sets.}
\begin{figure}[t]
\begin{center}
   \includegraphics[width=0.5\linewidth]{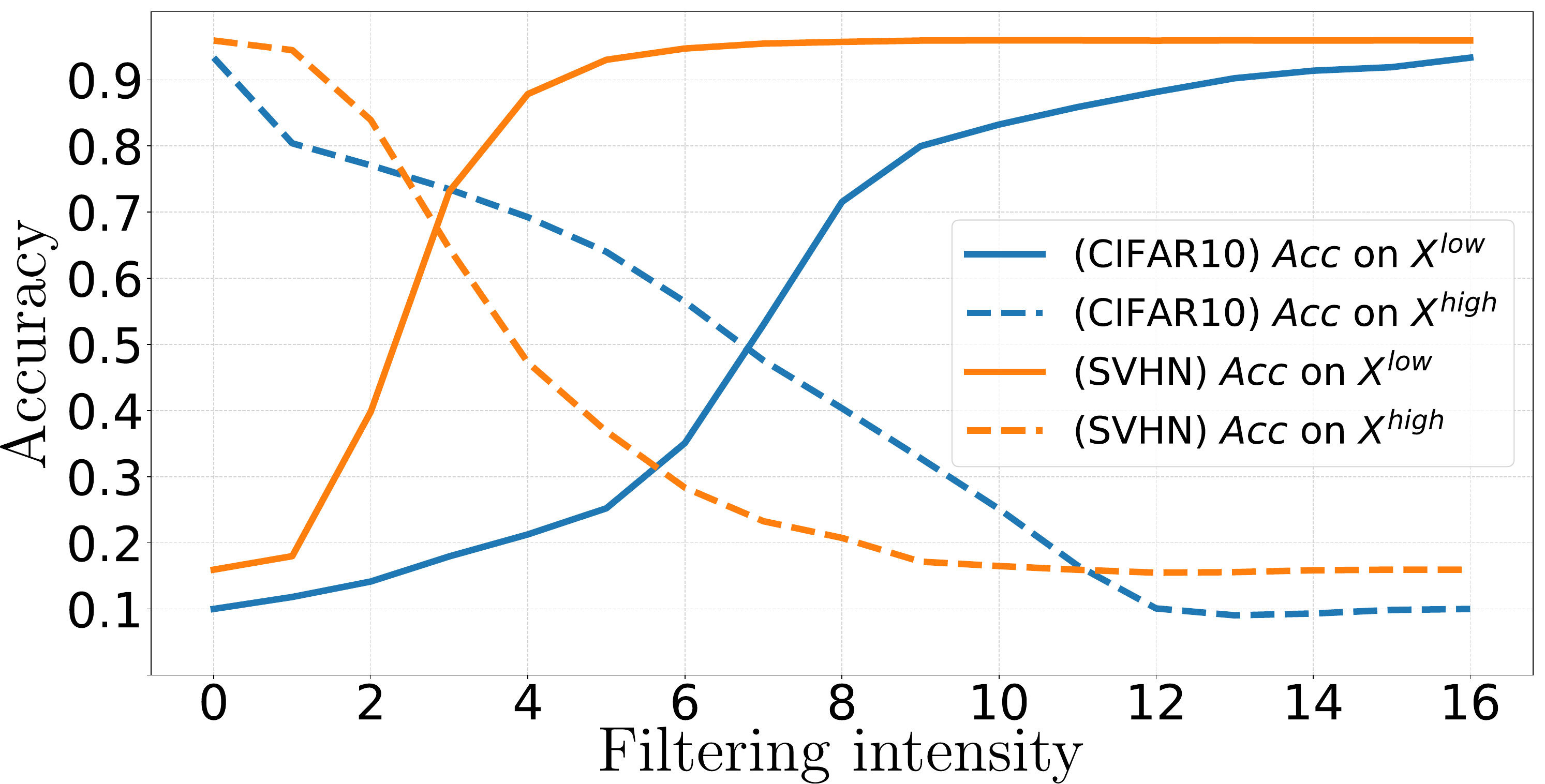}
\end{center}
\caption{CIFAR10 and SVHN. Accuracy of a regular model on low-pass  and high-pass filtered data set, for different filtering intensities.}
\label{cifar10_svhn_acc}
\end{figure}

\begin{figure}[t]
\begin{center}
   \includegraphics[width=0.45\linewidth]{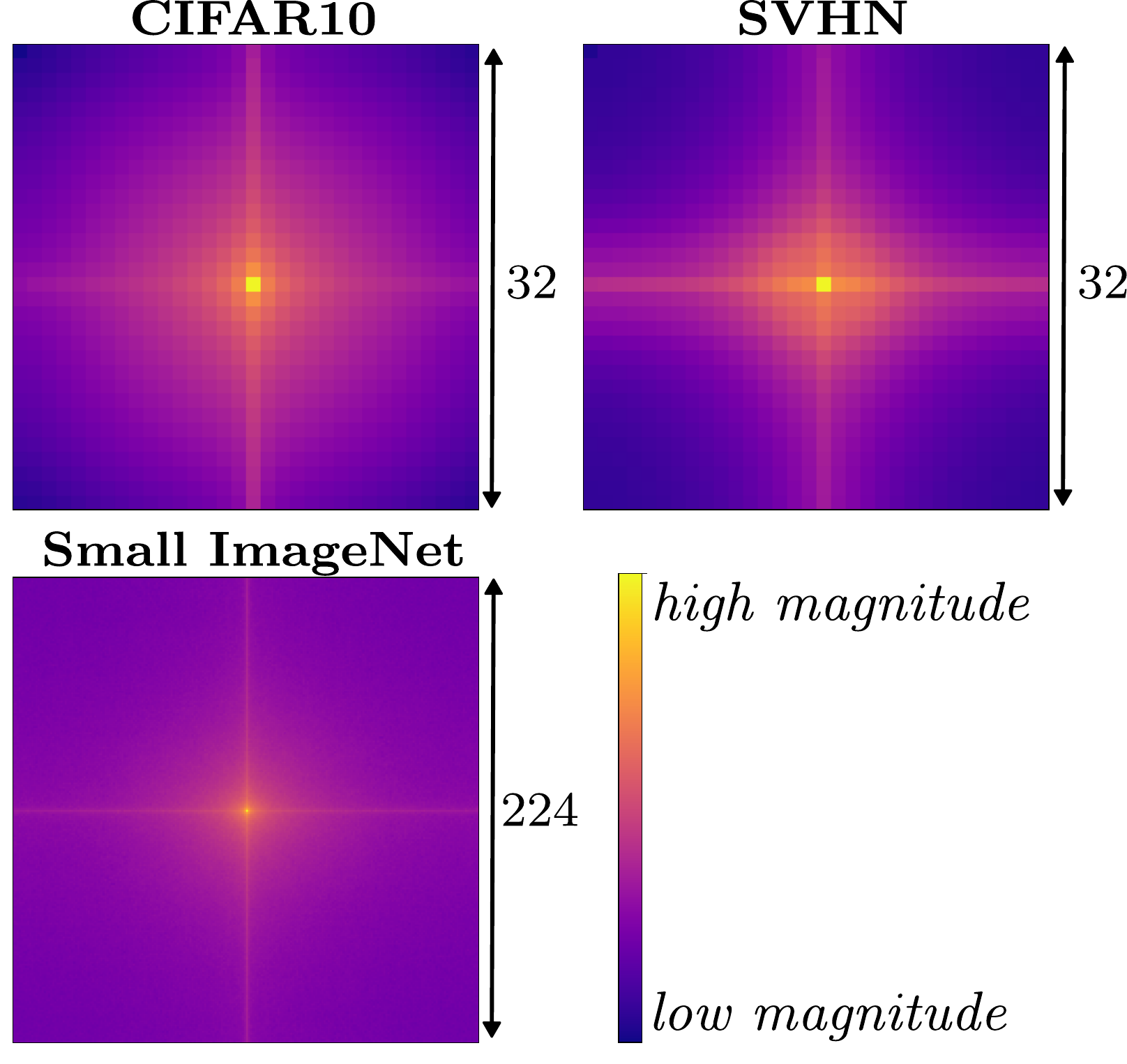}
\end{center}
\caption{CIFAR10, SVHN and Small Imagenet data sets.  Magnitude of the Fourier spectrum for clean images. Low frequencies are at the center, and high frequencies at the corners.}
\label{cifar10_svhn_imnet_freq}
\end{figure}

We begin by evaluating CIFAR10 and SVHN regular models on low-pass and high-pass filtered images. Results are presented in Figure \ref{cifar10_svhn_acc}. For CIFAR10, we notice that the accuracy of a model decreases much slower than SVHN when evaluated on more and more high-pass filtered images. We obtain an opposite result for SVHN with low-pass filtered data.
We can therefore assume that the informative features learned by the regular model are more focused on the LSF task for SVHN and are more spread between LSF and HSF tasks for CIFAR10. This analysis is consistent with the intrinsic frequency features of the data sets that is revealed by a classical Fourier analysis (presented in~Figure \ref{cifar10_svhn_imnet_freq}, top row) that actually shows a quite narrow spectrum for SVHN (towards LSF) and a spread spectrum for CIFAR10. 

\begin{figure}[t]
\begin{center}
   \includegraphics[width=0.6\linewidth]{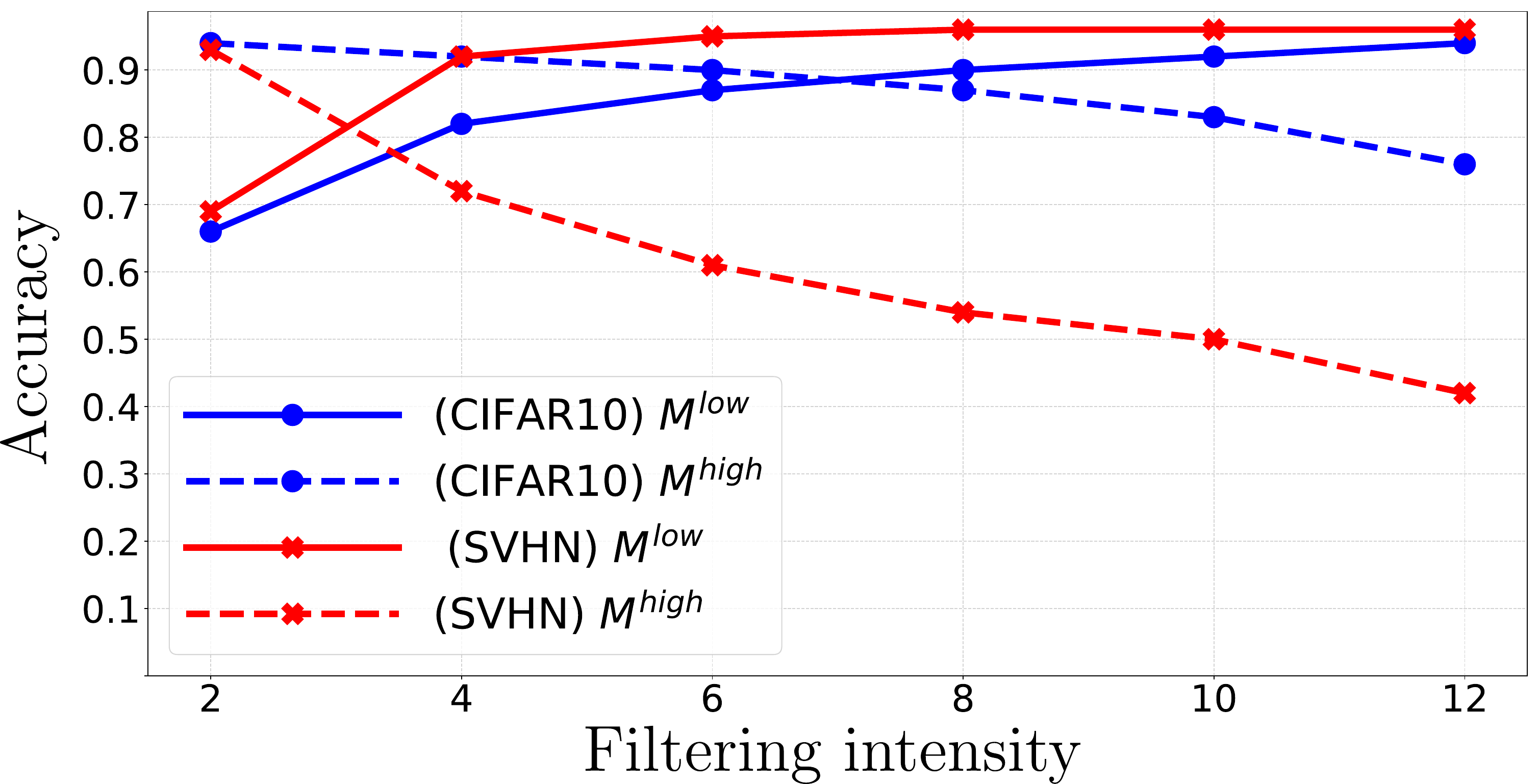}
\end{center}
\caption{CIFAR10 and SVHN. Test set accuracy of models trained on filtered data sets. (See Figure~\ref{filt_exemple} for filtering intensity effect)}
\label{cifar10_svhn_acc_tab}
\end{figure}

The accuracy of models $M^{low}_i$ and $M^{high}_i$ for various $i$, presented in Figure \ref{cifar10_svhn_acc_tab}, further highlights this phenomenon. The accuracy reached by models $M^{low}_i$ decreases much slower for SVHN than for CIFAR10, as the filtering intensity increases (and the inverse phenomenon is observable for high-pass filtering). This agrees with previous results (Figure \ref{cifar10_svhn_acc}), i.e. the useful information for the classification task is more distributed in the frequency spectrum for CIFAR10 compared to SVHN for which the information is predominantly concentrated in the low frequencies.

\subsection{Sensitivity to high spatial frequency noise.}

\begin{figure}[t!]
\begin{center}
   \includegraphics[width=0.4\linewidth]{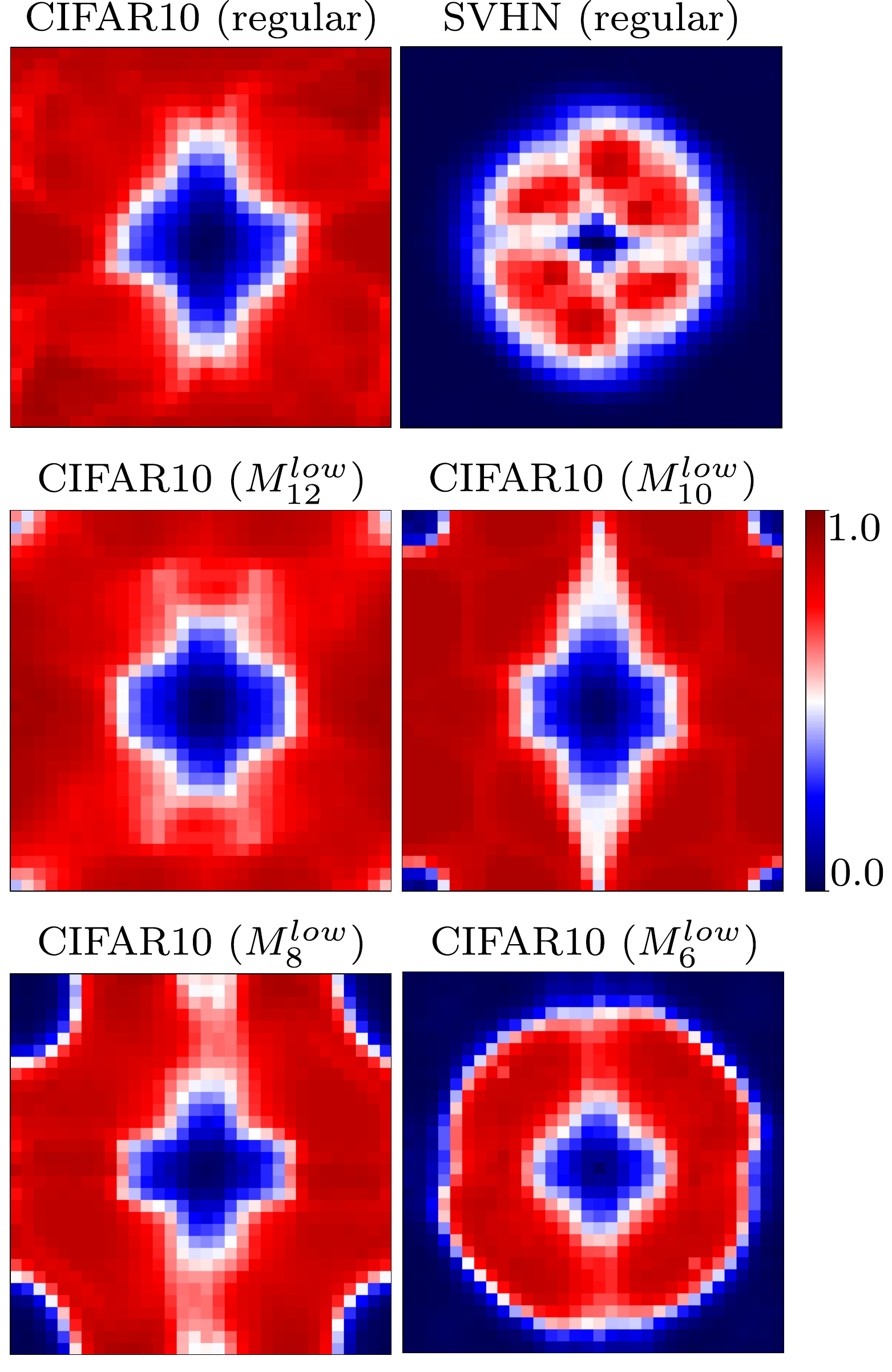}
\end{center}
\caption{CIFAR10 and SVHN. Error rate of models on data perturbed with Fourier constrained noise. Value at $(i,j)$ measures the error rate on data perturbed with noise along Fourier basis matrice $U_{i,j}$. Low frequencies are located at the center, and high frequencies at the corner. High values (red color) denote a high sensibility, and low values (blue color) denote a low sensibility.}
\label{cifar10_svhn_sensi_freq}
\end{figure}

We investigate the sensitivity of regular models to perturbations in specific frequencies. For that purpose, we use a similar procedure as in \cite{yinfourier2020}: the sensitivity is measured as the error rate of the model on a set of examples perturbed with noise located only in those spatial frequencies. More precisely, for a clean input image $x \in \mathbb{R}^{n \times n \times c}$, each channel is perturbed independently by the addition of a noise $rv U_{i,j}$, where $U_{i,j}$ is a Fourier basis matrix for coordinates $(i,j)$ in the Fourier domain, $r$ is chosen randomly in $\{ -1 ,1\}$, and $v$ controls the magnitude of the perturbation. We then measure the error rate of a model on $1,000$ well-classified test set examples perturbed with this type of noise, as a function of $(i,j)$. 
We focus our analysis on CIFAR10 and SVHN, two data sets with the same image size, and we experimentally set $v=4$. Results are presented in Figure \ref{cifar10_svhn_sensi_freq} (top row). We notice that a CIFAR10 regular model is more sensitive to LSF and HSF noise, while a SVHN regular model is more sensitive to LSF and not to high and very high frequencies.

For CIFAR10, we also investigate the sensitivity of models $M^{low}_i$. Results are presented in Figure \ref{cifar10_svhn_sensi_freq} (middle and bottom rows). Interestingly, we notice that the models $M^{low}_i$ become less sensitive to HSF as $i$ decreases (i.e. as the low-pass filtering intensity increases). Therefore, for CIFAR10, training a model on low-pass filtered images brings robustness against HSF perturbations. However, this robustness would be useless if adversarial perturbations were to rely on a broader spectrum than only HSF. We investigate this in the following section.

\label{sensitivity_noise}

\section{Transferability analysis}
\label{Transferability}
The transferability of adversarial perturbations between two models has been explained by shared non-robust useful features~\cite{ilyas2019adversarial}, that is features sensitive to adversarial perturbations and exploited for the prediction by both models. Therefore, to gain a better understanding of the nature of features at stake as well as frequency properties of adversarial perturbations, we study the transferability of adversarial examples between a regular model and models trained for the LSF or HSF task (i.e. $M^{low}$ and $M^{high}$). For these $M^{low}$ and $M^{high}$ models, a pre-processing layer is added before the model to evaluate it on non-filtered inputs. We use the $l_{\infty}$ DIM attack \cite{Xie2018ImprovingTO} (a state-of-the-art gradient-based attack tuned for transferability), with 40 iterations, a probability $p=0.8$ and a $l_{\infty}$ perturbation budget of $0.03$. 

\begin{figure}[t]
\begin{center}
   \includegraphics[width=0.6\linewidth]{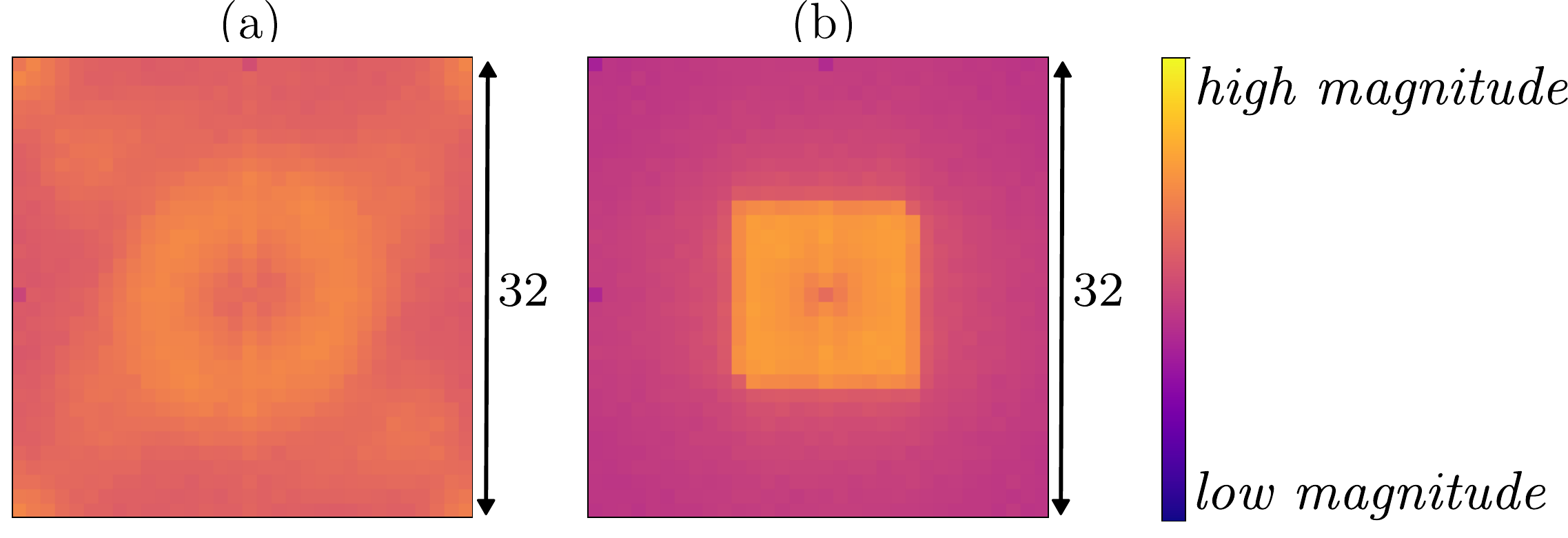}
\end{center}
\caption{CIFAR10. Magnitude of the Fourier spectrum for the adversarial perturbation  (a) Regular model, (b) Model trained on low-pass filtered dataset at intensity $6$}
\label{cifar10_adversarial_perturbation}
\end{figure}

We first remind that for models $M^{low}_i$, a low filtering intensity $i$ means that the model is trained on a strict low-pass filtered data set (see Figure~\ref{filt_exemple}) and as the filtering intensity $i$ increases, more and more HSF are considered. Thus, for example, adversarial examples crafted on $M^{low}_2$ exploit non-robust features learned with only LSF information and adversarial examples crafted on $M^{low}_{12}$ will have the possibility to take advantage of non-robust features determined on a broader range of frequencies. On the contrary,  for models  $M^{high}_i$, a high filtering intensity $i$ means that only HSF are kept (see Figure~\ref{filt_exemple}), and a low intensity represents a larger spectrum, gathering more LSF. Here, adversarial examples crafted on $M^{high}_{12}$ will exploit non-robust features learned exclusively on HSF information and adversarial examples crafted on $M^{high}_2$ will rely on 
non-robust features from a broader spectrum.
We illustrate this in Figure~\ref{cifar10_adversarial_perturbation} \textit{(a)} and \textit{(b)}, by presenting the magnitude of the Fourier spectrum of adversarial perturbation crafted on the regular model and on $M^{low}_6$. We observe that the perturbation is quite uniformly distributed along the spectrum (confirming a result from~\cite{yinfourier2020}) for the regular model and predominantly focused on the low frequencies for $M^{low}_6$. This is an important point, as it shows that adversarial examples are not HSF phenomena but may rely on a large range of spatial frequencies to efficiently fool a model. 

\begin{figure*}[b]
\begin{center}
   \includegraphics[width=\linewidth]{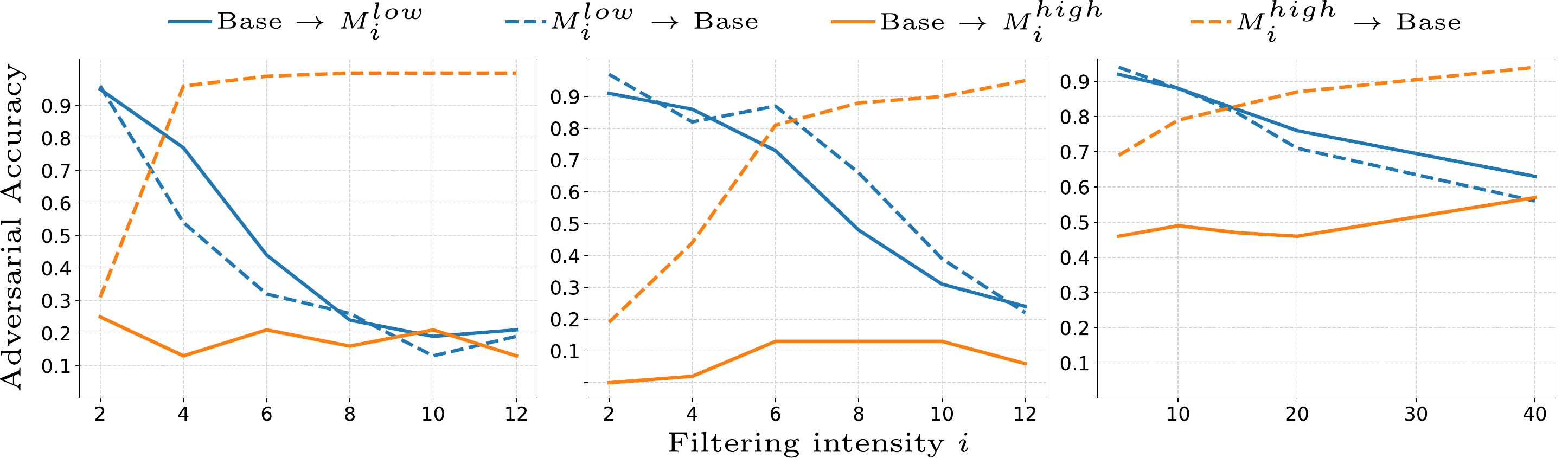}
\end{center}
\caption{SVHN (left), CIFAR10 (middle), Small ImageNet (right). Transferability analysis between base model and models trained on a filtered data set. (See Figure~\ref{filt_exemple} for filtering intensity effect)}
\label{cifar10_transferability}
\end{figure*}

Furthermore, analyzing the transferability results of Figure~\ref{cifar10_transferability}, we can provide interesting information about the nature of the robust and non robust features of a model as well as the properties of adversarial perturbations. Importantly, we observe similar behaviors between CIFAR10, SVHN and Small ImageNet.

A first conclusion comes from the two-way transferability between the regular model $M$ (note \textit{Base}) and models $M^{low}_i$ (blue curves). Indeed, we notice that the stronger the low-pass filtering (smaller $i$), the lower the transferability of adversarial perturbations.
This indicates that the regular classification task and the LSF task share predominantly robust useful features.

Secondly, an important outcome arises from the dissimilarity between the two orange curves for each data set. On one hand, the solid orange curve attests the impact of non-robust features exploiting HSF, highlighted by an almost constant success of adversarial examples crafted from the base model against models $M^{high}_i$ (about $0.1$, $0.2$ and $0.5$ of adversarial accuracy for CIFAR10, SVHN and Small ImageNet respectively).
On another hand, the dotted orange curve indicates that, as the high-pass filtering becomes more restrictive (i.e. as the $i$ value increases), the transferability of adversarial examples crafted on $M^{high}_i$ to the regular model decreases. These observations support the claim that,
to be efficient, adversarial perturbations must exploit a wide part of the spectrum, and therefore cannot be only focused on HSF.
This is particularly the case for Small ImageNet, as \textit{i) }the accuracy from a regular to a $M^{high}$ model is higher than for CIFAR10 and SVHN, and \textit{ii)} the transferability of adversarial examples crafted on $M^{high}_i$ is already poor for $i=5$.

We can summarize our conclusions as follows:
\begin{itemize}
\item Robustness is strongly related to features that rely on LSF information. 
\item Adversarial perturbation are not efficient when focused only on HSF.
\end{itemize}
In the next section, we propose to look deeper into the impact of frequency-based constraints when training a regular model in order to investigate if it can help increasing its robustness.

\section{Adversarial robustness of frequency-constrained models}
\label{Constraints on frequencies}
Following the precedent observations, we investigate loss functions designed to make a model to leverage informative features for both the regular task and the LSF and/or HSF task and evaluate how it impacts the adversarial robustness.
The adversarial robustness is evaluated with the adversarial accuracy ($Acc_{add}$) considering an attacker in the white-box setting, under the common $l_{\infty}$ threat model.
Adversarial examples are crafted with the $l_{\infty}$ PGD attack \cite{Madry2017}, with a perturbation budget of $\, \epsilon = 0.03$. To provide the more accurate evaluation of robustness as possible, notably to assess that no gradient masking occurs, we follow state-of-the-art guidelines from~\cite{carlini2019evaluating}. Detailed architectures and setups are presented in the code repository of this work.

    \subsection{Frequency-based regularization}
    \label{Constraint on low and high frequencies}
    
To enforce a model $M_{\theta}$ to rely on information relative to the LSF or HSF task, we define the following loss function $L^{freq}$, which acts on the logits. 

\begin{equation}
\begin{split}
    L^{freq}_{i,j}(\theta, x,y) = & L^{E}(\theta, x, y) 
      + \lambda_1 \left\| f(x) - f(x^{low}_i) \right\|_2^2  \\
    & + \lambda_2 \left\| f(x) - f(x^{high}_j) \right\|_2^2    
\end{split}
\label{loss L_all}
\end{equation}

For readability, when $\lambda_1=0$ (i.e. constraint is only focused on the HSF task) the loss is simply noted as $L^{high}_i$, and for $\lambda_2=0$ as $L^{low}_i$. During training, the cross-entropy part $L^{E}$ makes the model to learn useful features to solve the regular classification task. The second part (moderated by $\lambda_1$ and $\lambda_2$) constraints the model to extract useful features coherently to the LSF and HSF tasks. 

For CIFAR10 and SVHN, the loss functions considered are $L^{low}_i$ and $L^{high}_i$ for $i \in \left\{2,4,6,8,10\right\}$, and $L^{freq}_{i,j}$
for $(i,j) \in \left\{ (10,4), (4,12), (6,3), (6,10), (8,6), (6,8) \right\}$. For Small ImageNet, as each model training is costly, we consider the loss functions $L^{low}_i$, $L^{high}_i$  and $L^{freq}_{i,j}$ for representatives values in $\left\{ 10,20,40,60 \right\}$.
For conciseness purpose when presenting results, the subscript '${*}$' means that equal results are reached whatever the intensity of the filtering.

\subsection{Do the intrinsic frequency properties of the data bias the level of adversarial robustness?}
\label{iv_b}
\begin{table}[t]

\begin{center}
\resizebox{0.5\linewidth}{!}{
\begin{tabular}{c|cc|cc}

\multicolumn{5}{c}{\textbf{SVHN}}\\
\hline
& \textit{$Acc$} & \textit{$Constraints$} & \textit{$Acc_{adv}$} & \textit{$Constraints$} \\  
\hline
Best & 0.96 & $L^{low}_{i>2}$ & 0.41 & $L^{low}_{6}$\\
Worst & 0.93 & $L^{low}_{2}$ & 0.0 & $L^{low}_{10}$\\

\hline
\multicolumn{5}{c}{\textbf{CIFAR10}}\\
\hline
& \textit{$Acc$} & \textit{$Constraints$} & \textit{$Acc_{adv}$} & \textit{$Constraints$} \\  
\hline
Best & 0.92 & $L^{low}_2$, $L^{high}_{10}$ & 0.0 & $L^{low}_{*}$, $L^{high}_{*}$\\
Worst & 0.74 & $L^{low}_{2}$ & 0.0 & $L^{low}_{*}$, $L^{high}_{*}$\\

\hline
\multicolumn{5}{c}{\textbf{Small Imagenet}}\\
\hline
& \textit{$Acc$} & \textit{$Constraints$} & \textit{$Acc_{adv}$} & \textit{$Constraints$} \\  
\hline
Best & 0.92 & $L^{low}_{40}$ & 0.0 & $L^{low}_{*}$, $L^{high}_{*}$\\
Worst & 0.88 & $L^{high}_{40}$ & 0.0 & $L^{low}_{*}$, $L^{high}_{*}$\\
\hline

\end{tabular}
}
\end{center}
\caption{CIFAR10, SVHN and Small ImageNet. Best and worst accuracy on clean ($Acc$) and adversarial examples ($Acc_{adv}$) of models trained with the constrained loss $L^{low}_i$ or $L^{high}_i$.}
\label{cifar10_svhn_imnet_Llow_Lhigh}
\end{table}

Results for $L^{low}$ and $L^{high}$, presented in Table \ref{cifar10_svhn_imnet_Llow_Lhigh}, allow for a first important observation. Indeed, we see that the same constraint induces very different effects on the robustness, depending on the data set at stake. For CIFAR10 and Small ImageNet we observe no robustness when considering separate losses $L^{low}_{*}$ or $L^{high}_{*}$. On the contrary, models trained on SVHN with $L^{low}$ present an interesting level of adversarial robustness depending on the intensity level.
This observation appears as coherent with different frequency properties of the data sets highlighted in Section \ref{Frequency analysis and transferability}: the information learned by models trained on CIFAR10 and Small ImageNet are spread over the whole frequency spectrum, and -- on the contrary -- focused on LSF for SVHN. To further investigate this link, we proceed to check if a CIFAR10 model trained with $L^{low}$ would still reach some robustness on low-pass filtered data (i.e. train with $L^{low}_{i}(\theta,x^{low},y)$). In other words, we try to exclude the high spatial frequencies, which are predominantly non robust, and that we assume to explain this complete lack of robustness. To that purpose, we train models with $L^{low}_i$ ($i=6,8,10,12$) on low-pass filtered versions of CIFAR10 ($i=2,4,10$). Thus, we ensure that the model learns features strictly focused in the LSF.
Interestingly, we measure a slight but true and non-negligible robustness for some models with loss $L^{low}$ (up to an adversarial accuracy of $0.11$). 

As robustness is noticed for SVHN with $L^{low}$ or for CIFAR10 with $L^{low}$ on low-pass filtered data, a first conclusion is that a model relying predominantly on useful features of the LSF task can be made more robust with low-frequency based constraint ($L^{low}$).
Moreover, these models also share a non-sensitivity to high and very high frequencies (cf. Section~\ref{Frequency analysis and transferability}), which highlights another influential factor on the robustness of a model trained with the loss $L^{low}$. 

\begin{table}

\begin{center}
\resizebox{0.5\linewidth}{!}{
\begin{tabular}{c|p{0.3cm}p{2.4cm}|p{0.6cm}p{2.4cm}}
\multicolumn{5}{c}{\textbf{SVHN}}\\
\hline
& \textit{$Acc$} & \textit{$Constraints$} & \textit{$Acc_{adv}$} & \textit{$Constraints$} \\  
\hline
Best & 0.96 & $L^{freq}_{10,4}$, $L^{freq}_{6,*}$, $L^{freq}_{8,6}$ & 0.30 & $L^{freq}_{6,3}$\\
Worst & 0.95 & $L^{freq}_{4,12}$ & 0.0 & $L^{freq}_{10,4}$, $L^{freq}_{6,10}$, $L^{freq}_{6,8}$\\

\hline

\multicolumn{5}{c}{\textbf{CIFAR10}}\\
\hline
& \textit{$Acc$} & \textit{$Constraints$} & \textit{$Acc_{adv}$} & \textit{$Constraints$} \\  
\hline
Best & 0.95 & $L^{freq}_{5,3}$, $L^{freq}_{8,6}$ & 0.12 & $L^{freq}_{5,3}$, $L^{high}_{*}$\\
Worst & 0.92 & $L^{freq}_{4,12}$ & 0.03 & $L^{freq}_{4,12}$, $L^{freq}_{6,8}$\\
\hline

\multicolumn{5}{c}{\textbf{Small ImageNet}}\\
\hline
& \textit{$Acc$} & \textit{$Constraints$} & \textit{$Acc_{adv}$} & \textit{$Constraints$} \\  
\hline
Best & 0.92 & $L^{freq}_{60,*}$ & 0.36 & $L^{freq}_{40,20}$\\
Worst & 0.89 & $L^{freq}_{10,60}$ & 0.0 & $L^{freq}_{60,*}$\\
\hline
\end{tabular}
}
\end{center}
\caption{CIFAR10, SVHN and Small ImageNet. Best and worst accuracy on clean ($Acc$) and adversarial examples ($Acc_{adv}$) of models trained with the constrained loss $L^{freq}_{i,j}$.}
\label{cifar10_svhn_imnet_Lfreq}
\end{table}

However, is a frequency-based regularization suitable for complex data sets such as CIFAR10 or Small ImageNet? In Table \ref{cifar10_svhn_imnet_Lfreq}, we present results with a constraint spanning a wider spectrum ($\lambda_1 > 0$, $\lambda_2 > 0$). For SVHN, as expected, the combined constraints does not enable to reach better robustness compared to the loss $L^{low}$, as the combination of the two constraints is not compatible with the intrinsic frequency properties of the data set: information is predominantly concentrated in the low frequencies. For Small ImageNet and CIFAR10 the combined constraint forces the model to learn informative features for the LSF and HSF tasks, which are mainly robust as features informative of the LSF task are mostly robust, as shown in Section~\ref{Transferability} for CIFAR10. Interestingly, stronger level of robustness is observed for Small ImageNet and we hypothesize that the impact of the combined constraint is all the more efficient that the frequency spectrum is wider.

\subsection{Is frequency-based regularization compatible with adversarial training ?}
\label{Combination with adversarial training}
\begin{table}
\begin{center}
\resizebox{0.4\linewidth}{!}{
\begin{tabular}{c|c|c|c|c}
\multicolumn{5}{c}{Loss $L_{AT}$: $Acc = 0.84$, $Ac_{adv}=0.55$}\\
\hline
\multicolumn{5}{c}{Loss $L^{AT,low}_i$}\\
\hline
$i=$ & \textit{2} & \textit{4} & \textit{8} & \textit{10}\\
\hline
$Acc$        & 0.80 & 0.79 & 0.81 & 0.82 \\
$Acc_{adv}$  & 0.59 & 0.60 & 0.57 & 0.56 \\
\hline
\multicolumn{5}{c}{Loss $L^{AT,high}_i$}\\
\hline
$i=$ & \textit{4} & \textit{\textbf{6}} & \textit{8} & \textit{10} \\ 
\hline
$Acc$       & 0.84 & \textbf{0.84} & 0.85 & 0.83 \\
$Acc_{adv}$ & 0.58 & \textbf{0.6} & 0.55 & 0.60 \\
\hline
\multicolumn{5}{c}{Loss $L^{AT,freq}_{i,j}$} \\
\hline
$(i,j)=$ & \textit{(10,4)} & \textit{(2,10)} & \textit{(4,8)} & \textit{(8,6)} \\ 
\hline
$Acc$       & 0.84 & 0.8 & 0.81 & 0.82 \\
$Acc_{adv}$ & 0.58 & 0.61 & 0.58 & 0.6 \\
\hline
\end{tabular}
}
\end{center}
\caption{CIFAR10. Accuracy on clean ($Acc$) and adversarial ($Acc_{adv}$) examples of models trained with loss functions $L^{AT,low}_i$ ($\lambda_2=0$), $L^{AT,high}_i$ ($\lambda_1=0$) and $L^{AT,freq}_{i,j}$}
\label{cifar10_adv_t}
\end{table}
\begin{table}[t]
\begin{center}
\resizebox{0.40\linewidth}{!}{
\begin{tabular}{c|c|c|c|c}
\multicolumn{5}{c}{Loss $L_{AT}$: $Acc = 0.93$, $Ac_{adv}=0.54$}\\
\hline
\multicolumn{5}{c}{Loss $L^{AT,low}_i$}\\
\hline
$i=$ & \textit{2} & \textit{4} & \textit{8} & \textit{10}\\
\hline
$Acc$       & 0.87 & 0.9 & 0.90 & 0.90 \\
$Acc_{adv}$ & 0.59 & 0.59 & 0.60 & 0.61 \\
\hline
\multicolumn{5}{c}{Loss $L^{AT,high}_i$}\\
\hline
$i=$ & \textit{4} & \textit{6} & \textit{8} & \textit{10} \\  
\hline
$Acc$       & 0.87 & 0.8 & 0.88 & 0.87 \\
$Acc_{adv}$ & 0.56 & 0.58 & 0.57 & 0.57 \\
\hline
\multicolumn{5}{c}{Loss $L^{AT,freq}_{i,j}$} \\
\hline
$(i,j)=$ & \textit{\textbf{(10,4)}} & \textit{(2,10)} & \textit{(4,8)} & \textit{(8,6)} \\
\hline
$Acc$       & \textbf{0.92} & 0.92 & 0.91 & 0.91 \\
$Acc_{adv}$ & \textbf{0.66} & 0.6 & 0.64 & 0.66 \\
\hline
\end{tabular}
}
\end{center}
\caption{SVHN. Accuracy on clean ($Acc$) and adversarial ($Acc_{adv}$) examples of models trained with loss functions $L^{AT,low}_i$ ($\lambda_2=0$), $L^{AT,high}_i$ ($\lambda_1=0$) and $L^{AT,freq}_{i,j}$}
\label{svhn_adv_t}
\vspace{-0.4cm}
\end{table}

Motivated by the true robustness observed when training some models with loss functions $L^{low}$, $L^{high}$ or $L^{freq}$, we study the combination of these losses with  Adversarial Training (hereafter AT) \cite{Madry2017}, a common and widely used approach when defending against an adversary in the white-box setting. This study is of particular interest as AT is shown to be related to frequency concern as it makes a model rely more on low-frequency concepts and being less sensitive to perturbations in the high-frequencies \cite{Zhang2019interpreting, geirhos2018imagenettrained}. AT consists of training a model on adversarial examples generated online and designed to induce a worst-case loss. Considering a $l_{\infty}$ bound $\epsilon$ for adversarial perturbations and an input-label pair $(x,y)$, an adversarial example $x' = x + \delta $ is generated following:
\begin{equation}
    \delta = \argmax_{\left\| \delta \right\|_{\infty} \, \leq \, \epsilon} \quad L^{E}(\theta, x + \delta ,y)
\end{equation}
The model is then trained minimizing  the loss $L_{AT} = L^{E}(\theta,x + \delta,y)$. For the combination of Adversarial Training with frequency-based constraints, we define the loss functions $L^{AT, freq}$:

\begin{equation}
 L^{AT,freq}_{i,j} (\theta, x,y) = L^{freq}_{i,j}(\theta, x + \delta,y)
\end{equation}

As AT can be prone to robust overfitting \cite{rice2020overfitting},
we perform early-stopping relatively to the error on an hold-out set of test set examples when training models with these loss functions.
We evaluate the robustness against the $l_{\infty}$ PGD attack with $\epsilon=0.03$ for CIFAR10 and SVHN, considering the same sanity checks and attack parameters as the ones used in Section \ref{Constraints on frequencies}, to ensure the tightest evaluation of robustness as possible.
Results are presented in Tables \ref{cifar10_adv_t} and \ref{svhn_adv_t} for CIFAR10 and SVHN, respectively. 
We highlight the fact that, when dealing with Adversarial Training, there is an inherent trade-off between accuracy and robustness \cite{zhang2019theoretically}. Therefore, for a fair comparison, we focus on the benefits of the addition of frequency-based constraints, only for no (or up to $0.1 \%$) loss of natural accuracy (highlighted in bold in Tables \ref{cifar10_adv_t} and \ref{svhn_adv_t}). Particularly, for CIFAR10, a model trained with loss $L_{AT,all,2,10}$ outperforms AT by $0.06$ on adversarial accuracy. For SVHN, this improvement reaches $0.12$ for a model trained with loss $L_{AT,all,10,4}$.

From these results, two important and dependent observations can be made.
In first place, it shows that existing defense scheme can in fact benefit from constraints related to frequency properties.
However, in second place, it shows that the intrinsic effects of the defense scheme (AT) has an influence on the effectiveness of such frequency-based regularization.
Indeed, relatively to each data set, there is no connection between the frequency constraints which allow the loss function $L^{AT,freq}$ to outperform Adversarial Training, and the ones which bring robustness when considered alone (\ref{iv_b}). Indeed, as an example, for CIFAR10, no robustness was brought by constraint on the HSF task (Table \ref{cifar10_svhn_imnet_Llow_Lhigh}), whereas this is the loss $L^{AT, high}_i$ which outperforms Adversarial Training. A possible explanation is very likely to come from Adversarial Training biasing the sensitivity of models towards the low frequencies. 
Therefore, when combining frequency related constraint with a defense scheme, not only do the frequency properties of the data at stake play an important role, but frequency related effects of the defense itself has a strong influence.

\section{Conclusion}
\label{Conclusion}

In this paper, we investigate through experiments the link between frequency-based processing and adversarial robustness. 
Particularly, this allows to gain insight on the strong influence of frequency properties of the data at stake that may be very specific according to the application domain. 
Notably, we found that models relying predominantly on useful features for the LSF task, and with a non-sensitivity to high frequency noise show robustness when constrained to rely on useful information for the LSF task.
Interestingly, when the information encompassed in the images is spread over the whole frequency spectrum, a constraint spanning a wide frequency spectrum is a viable solution.
Moreover, the efficiency of frequency-based regularization when combined with existing defense schemes is strongly dependent of the nature of these schemes.
These experiences as well as the conclusions of previous efforts in neural computation and cognitive psychology highlight the fact that the intrinsic frequency characteristics of data must be necessarily considered when designing robust defense strategies against integrity-based attacks of supervised models. 

\section*{Acknowledgments}
This work is a collaborative action that is partially supported by the European project ECSEL InSecTT\footnote{\url{www.insectt.eu}, InSecTT: ECSEL Joint Undertaking (JU) under grant agreement No 876038. The JU receives support from the European Union’s Horizon 2020 research and innovation program and Austria, Sweden, Spain, Italy, France, Portugal, Ireland, Finland, Slovenia, Poland, Netherlands, Turkey. The document reflects only the author’s view and the Commission is not responsible for any use that may be made of the information it contains.} and by the French National Research Agency (ANR) in the framework of the \textit{Investissements d’avenir} program (ANR-10-AIRT-05, irtnanoelec)
~and benefited from the French Jean Zay supercomputer thanks to the \textit{AI dynamic access} program.

\bibliographystyle{ieeetr}
\bibliography{Biblio.bib}

\end{document}